\documentclass[twocolumn]{IEEEtran}

\usepackage[pdftex]{graphicx} 
\usepackage{amsmath} 
\usepackage{algorithmic}
\usepackage{array} 
\usepackage[caption=false,font=footnotesize]{subfig} 
\usepackage{dblfloatfix}
\usepackage{url} 

\usepackage{microtype}
\usepackage[utf8]{inputenc} 
\usepackage[T1]{fontenc}    
\usepackage{paralist} 
\usepackage{marvosym} 
\usepackage{bm} 
\usepackage{bbm} 
\usepackage{amssymb}   
\usepackage{mathrsfs} 
\usepackage{tabularray} 
\UseTblrLibrary{booktabs}       
\usepackage{tabularx} 
\usepackage[flushleft]{threeparttable} 
\usepackage{colortbl} 
\usepackage{makecell} 
\usepackage{multirow} 
\usepackage{nicefrac} 
\usepackage[dvipsnames]{xcolor} 
\usepackage{soul} 
\setul{0.5ex}{0.3ex}

\definecolor{citecolor}{RGB}{34,139,34}
\usepackage[pagebackref=false,breaklinks=true,letterpaper=true,colorlinks,
    citecolor=citecolor,bookmarks=false]{hyperref}
\usepackage{cleveref}  
\crefformat{figure}{Fig.~#2#1#3}
\crefformat{equation}{Eq.~(#2#1#3)}
\crefformat{table}{Table~#2#1#3}
\crefformat{section}{Section~#2#1#3}
\crefformat{algorithm}{Algorithm~#2#1#3}
\crefformat{appendix}{Appendix #2#1#3}
\crefformat{subsection}{subsection~#2#1#3}
\usepackage{cellspace}
\usepackage[numbers,sort&compress]{natbib} 
\usepackage{wrapfig}
\usepackage{xspace} 

\usepackage{pifont} 
\newcommand{\xmark}{\ding{53}}%

\definecolor{Gray}{rgb}{0.9,0.9,0.9}
\definecolor{LightCyan}{rgb}{0.88,1,1}
\newcolumntype{a}{>{\columncolor{Gray}}c}
\newcolumntype{b}{>{\columncolor{white}}c}

\begin{document}

\setlength{\abovedisplayskip}{.5\baselineskip} 
\setlength{\belowdisplayskip}{.5\baselineskip} 

\title{Background Semantics Matter: Cross-Task Feature Exchange Network for Clustered Infrared \\Small Target Detection}



\author{

  Mengxuan~Xiao,
  Yinfei~Zhu,
  Yiming~Zhu,
  Boyang Li,
  Feifei Zhang,
  Huan~Wang,
  Meng~Cai,
  Yimian~Dai
  \thanks{
    This work was supported by
      the National Natural Science Foundation of China (No. 62301261, 
        62472443). 
    \emph{The first two authors contributed equally to this work. (Corresponding author:
      Yiming Zhu, Huan Wang, and Yimian Dai).}
    }

  \thanks{
    Mengxuan Xiao, 
    Yiming Zhu and Huan Wang
    are with School of Computer Science and Engineering, Nanjing University of Science and Technology, Nanjing, China.
    (e-mail:
    \href{mailto:xiaomengxuan@njust.edu.cn}{xiaomengxuan@njust.edu.cn};
    \href{mailto:yiming\_zhu\_grokcv@163.com}{yiming\_zhu\_grokcv@163.com};
    \href{mailto:wanghuanphd@njust.edu.cn}{wanghuanphd@njust.edu.cn}).
  }

  \thanks{
    Yinfei Zhu and Meng Cai are with the Luoyang Institute of Electro-Optical Equipment under AVIC (Aviation Industry Corporation of China).
    \href{mailto:zhuyf\_613@163.com}{zhuyf\_613@163.com}.
  }
  
  \thanks{
    Boyang Li is with College of Electronic Science and Technology, National University of Defense Technology, Changsha, China.
    (e-mail:
    \href{mailto:xiaomengxuan@njust.edu.cn}{liboyang20@nudt.edu.cn}.
  }
  

  %
  \thanks{
  Feifei Zhang is with School of Computer Science, Guangdong University of Education. 
    (e-mail: \href{mailto:zhangfeifei2006@126.com}{zhangfeifei2006@126.com}).
  }


  \thanks{
  Yimian Dai is with VCIP, College of Computer Science, Nankai University. (e-mail:
  \href{mailto:yimian.dai@gmail.com}{yimian.dai@gmail.com})
  }
  
}

\maketitle


\begin{abstract}

Infrared small target detection presents significant challenges due to the limited intrinsic features of the target and the overwhelming presence of visually similar background distractors. We contend that background semantics are critical for distinguishing between objects that appear visually similar in this context. To address this challenge, we propose a task—clustered infrared small target detection—and introduce DenseSIRST, a benchmark dataset that provides per-pixel semantic annotations for background regions. This dataset facilitates the shift from sparse to dense target detection. Building on this resource, we propose the Background-Aware Feature Exchange Network (BAFE-Net), a multi-task architecture that jointly tackles target detection and background semantic segmentation. BAFE-Net incorporates a dynamic cross-task feature hard-exchange mechanism, enabling the effective exchange of target and background semantics between the two tasks. Comprehensive experiments demonstrate that BAFE-Net significantly enhances target detection accuracy while mitigating false alarms. The DenseSIRST dataset, along with the code and trained models, is publicly available at \url{https://github.com/GrokCV/BAFE-Net}.

\end{abstract}


\section{Introduction} \label{sec:introduction}

\begin{figure}[hbtp]
\vspace{-1\baselineskip}
  \centering
  \includegraphics[width=0.98\linewidth]{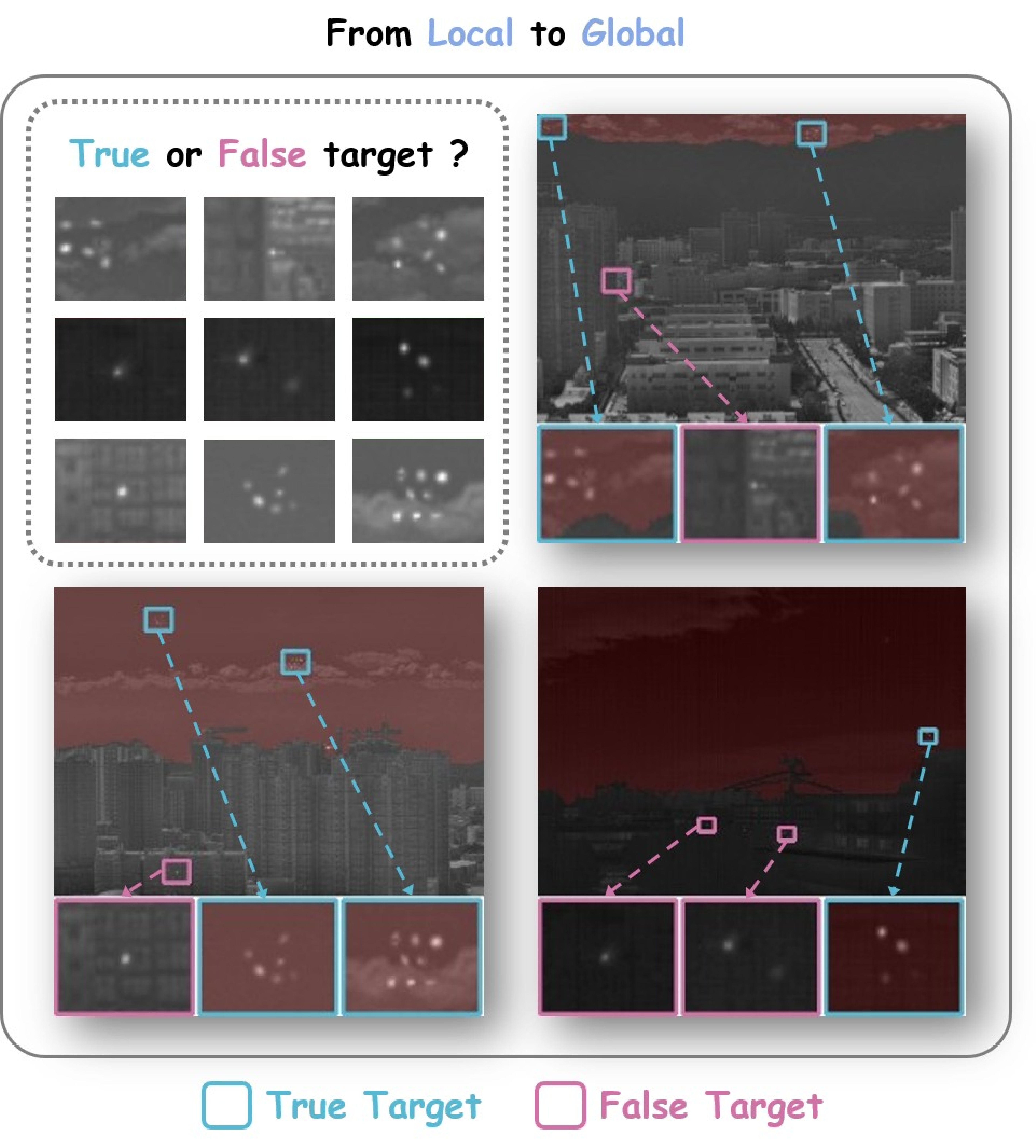}
  \vspace{-0.5\baselineskip}
  \caption{
  \textbf{Illustration of the background semantics' crucial role in infrared small target detection.} When considering only local regions (as in the top-left image), it is challenging to differentiate between clustered small targets and false alarms due to their similar appearances. By incorporating global contextual information (as in the surrounding images), the distinction between real targets and false alarms becomes more apparent.
  From the global images, it can be seen that the sky background is highlighted in red. The small target in the blue box represents a genuine target of interest against the sky background, while the small target in the red box is a false alarm in the context of the building background.
  }
  \label{fig:difficulty}
\vspace{-1\baselineskip}
\end{figure}

\IEEEPARstart{I}{nfrared} small target detection has emerged as a fundamental challenge in computer vision, with critical applications spanning military surveillance, drone detection, search and rescue missions, and night operations. The task involves identifying and localizing diminutive objects that appear as few pixels in infrared imagery, often captured under adverse weather conditions or limited visibility scenarios. As defense technologies evolve and multi-target scenarios become increasingly prevalent, the demand for robust and accurate detection systems has intensified dramatically.

\subsection{Challenges on Clustered Infrared Small Target Detection}

Traditional infrared small target detection research has predominantly focused on sparse target scenarios~\cite{gao2013infrared, dai2017reweighted}, where isolated objects are distributed across relatively uniform backgrounds. However, contemporary real-world applications present fundamentally different challenges. In defense against drone swarms, multi-target tracking systems, and large-scale maritime rescue operations, targets frequently appear in clustered configurations, creating dense distributions that fundamentally alter the detection paradigm.
This gap underscores an urgent need for advanced technology that can effectively identify and track \textbf{\textit{clustered infrared small targets}}.


This transition from sparse to dense target detection introduces a cascade of technical challenges that render existing methodologies inadequate:


\begin{enumerate}
    \item Methods that rely on the assumption of sparsely distributed targets become inapplicable in dense scenes, where targets are closely situated and may occlude one another, leading to the breakdown of single-target analysis strategies.
    
    \item Local contrast based methods~\cite{chen2013local}, which depend on detecting objects against uniform backgrounds, become unreliable as dense clusters of targets blur the line between target and background.
    
    \item Even state-of-the-art deep learning methods~\cite{zhu2024towards, zhang2022isnet, dai2022ao2} struggle when confronted with dense target scenarios where the delineation between targets and background blurs, adding to the complexity of detection tasks. 

\end{enumerate}

These challenges are compounded by the limitations in current datasets, which primarily focus on sparse target detection and fail to address the complexities of clustered targets.

\subsection{Limitations of Current Approaches}

While the rapid advancements in generic object detection have led to state-of-the-art methods like TOOD~\cite{feng2021tood} and CrossKD~\cite{wang2023crosskd} that address dense object detection, these techniques cannot be directly applied to the unique challenges posed by infrared small target detection. The specialized nature of this field presents two critical factors that hinder the direct transp lantation of generic vision approaches:
\begin{enumerate}
    \item \textbf{Scarcity of Target-Specific Features:}
    Infrared small targets typically occupy only a few pixels and lack distinctive visual characteristics such as texture, shape, or structural details. This ``visual poverty'' renders traditional feature based recognition and localization methods largely ineffective, as there are insufficient discriminative cues to differentiate targets from background elements.
    \item \textbf{Abundance of Similar Background Distractors:} 
    Infrared imagery frequently contains background elements~\cite{liao2022ccfenet}, such as thermal noise, hot spots, and structural artifacts, that exhibit visual similarity to genuine targets. This similarity creates a high false alarm rate when relying solely on local appearance-based discrimination.
\end{enumerate}
These challenges inevitably lead us to ponder a fundamental question: \textbf{\textit{What truly distinguishes visually similar objects in single-frame infrared small target detection, assigning them different class labels?}} Unraveling this enigma holds the key to developing effective and robust algorithms for this task.

\subsection{Motivation}

The answer to the aforementioned challenge lies in a paradigm shift: rather than relying solely on target appearance, we propose to exploit \textbf{\textit{background semantics}}.

Consider the scenario illustrated in Fig.~\ref{fig:difficulty}: when examining only local regions, distinguishing between genuine targets and false alarms becomes nearly impossible due to their visual similarity. However, by incorporating global contextual information, the classification task becomes significantly more tractable. A small bright object appearing against a sky background typically represents a target of interest, while an identical object against building infrastructure is more likely a false alarm. This contextual disambiguation demonstrates that background semantics\cite{Springer2024MRRFS} serve as crucial semantic anchors that enable accurate target identification where local appearance fails.

However, it is crucial to highlight a significant gap in the current landscape of infrared small target detection research. Despite the availability of several public datasets~\cite{dai2021asymmetric,dai2023one,li2022dense,zhang2022isnet},
\textbf{\textit{none of these datasets provide detailed semantic annotations for the background regions}}. 

Without these annotations, it is challenging to explicitly model the background semantics necessary for improving detection performance. To bridge this critical gap and validate our hypothesis, we introduce DenseSIRST. This is a comprehensive infrared target detection benchmark that addresses both the clustered target detection challenge and the background semantic modeling requirement. Our dataset implements two revolutionary improvements over existing resources:

\begin{itemize}
    \item Dense Target Scenarios: Transitioning from sparse to clustered target configurations that reflect real-world deployment conditions such as drone swarm detection and multi-target tracking applications.
    \item Background Semantic Annotations: Providing detailed per-pixel semantic segmentation of background regions, categorized into sky and non-sky classes, thereby enabling explicit background semantic modeling for the first time in this domain.
\end{itemize}


Building upon the semantic-rich DenseSIRST dataset, we propose the Background-Aware Feature Exchange Network(BAFE-Net) framework, which fundamentally transforms infrared small target detection from a single target detection task that solely focuses on the foreground to a multi-task architecture that jointly performs target detection and background semantic segmentation. 
The core innovation of BAFE-Net lies in its \textbf{dynamic cross-task feature hard-exchange module}, which facilitates selective semantic information transfer between detection and segmentation tasks through an adaptive feature adapter. \textit{This represents the first implementation of cross-task feature hard-exchange mechanisms in infrared target detection}. By explicitly modeling background semantics, BAFE-Net achieves superior contextual understanding, resulting in significant improvements in detection accuracy while substantially reducing false alarm rates.

In summary, our contributions can be categorized into three main aspects:
\begin{enumerate}
    \item We propose a new task—-clustered infrared small target detection, which poses unique challenges compared to traditional single-target detection.
    \item We have released a new dataset, DenseSIRST, annotated with background semantic segmentation. This dataset offers rich contextual information to enhance target detection.
    \item We propose the BAFE-Net, which achieves a dynamic mechanism of mutual promotion between segmentation and detection tasks by implementing selective feature hard-exchange with a cross-task adapter, based on explicit background modeling.
\end{enumerate}

\begin{table*}[b]
  \vspace{-1\baselineskip}
  \setlength{\abovecaptionskip}{0cm}  
  \renewcommand\arraystretch{1.2}
  \footnotesize
  \centering
  \vspace{-0.5\baselineskip}
  \caption{
  A Comparative Analysis of the DenseSIRST Dataset and Other Popular Datasets for Infrared Small Target Detection.
  }
  \label{tab:datasets}
  \setlength{\tabcolsep}{7.5pt}
  \begin{tabular}{l|c|c|c|c|c|c|c}
  \hline 
      Datasets & Image Type & Annotation Type & \thead{Image \\ Number} & \thead{Target \\ Number} & \thead{Average \\ Target Area} & \thead{Sparse or \\ Clustered} & \thead{Background Semantic \\ Annotation} \\
  \hline
      SIRST V1 \cite{dai2021asymmetric} & Real & Pixel & 427 & 533 & 23 & Sparse & \xmark \\
      SIRST V2 \cite{dai2023one} & Real & Pixel + BBox + Point & 1024 & 648 & 24 & Sparse & \xmark \\
      IRSTD1K \cite{zhang2022isnet} & Real & Pixel & 1001 & 1495 & 38 & Sparse & \xmark \\
      SIRSTAUG \cite{zhang2023attention} & Synthetic & Pixel & 8525 & 9278 & 88 & Sparse & \xmark \\
      \rowcolor[rgb]{0.9,0.9,0.9} \textbf{DenseSIRST (Ours)} & Real + Synthetic & Pixel + BBox + Point & 1024 & \textbf{13655} & \textbf{6} & \textbf{Cluster} & \checkmark \\
  \hline
\end{tabular}
\vspace{-0.5\baselineskip}
\end{table*}

\section{Related Work} \label{sec:related}

The core idea of traditional infrared small target detection algorithms is to suppress background interference and enhance target saliency. These methods mainly fall into three categories:
\begin{enumerate}
    \item \textit{Background Estimation Methods:} 
    By exploiting the local smoothness or non-local self-similarity of background regions, various filtering strategies are designed to highlight targets. However, these methods often fail when facing extremely weak or low-contrast targets, primarily due to their limited semantic perception capabilities and high sensitivity to handcrafted hyperparameters. Consequently, they are prone to both missed detections and false alarms in complex backgrounds.
    
    \item \textit{Human Visual System (HVS) Methods:} Techniques such as local contrast measure (LCM) \cite{chen2013local} and multi-scale patch-based contrast measure (MPCM) \cite{wei2016multiscale} attempt to emulate the high spatial local contrast sensitivity of the human eye. 
    While intuitive and computationally efficient, their reliance on pixel-level or patch-level contrast leads to diminished performance when the target is extremely small, blurred, or embedded in cluttered textures.
    
    \item \textit{Low-Rank and Sparse Decomposition Methods:} 
    Inspired by matrix and tensor decomposition theories, methods such as the Infrared Patch-Image (IPI) model~\cite{gao2013infrared}, the Reweighted Infrared Patch-Tensor (RIPT) model~\cite{dai2017reweighted} and the low-rank tensor approximation with saliency prior (LRTA-SP) model~\cite{TAES2024LRTASP} decompose the image into low-rank background and sparse target components. Despite demonstrating strong background suppression capabilities, they struggle to differentiate true targets from structured background noise or distractors due to the shared sparsity characteristics, limiting their robustness in real-world scenarios.
\end{enumerate}

With the advent of deep learning, data-driven methods have significantly advanced infrared small target detection by learning robust representations directly from data, rather than relying on handcrafted priors. These approaches leverage deep neural networks to automatically extract semantic features, enabling better adaptation to diverse and complex backgrounds.

Several representative methods have demonstrated the advantages of deep architectures in modeling contextual information and capturing multi-scale cues. For example, the asymmetric context modulation network (ACM)~\cite{dai2021asymmetric} introduces a context-sensitive structure to enhance the discrimination between targets and background. DNANet~\cite{li2022dense} further incorporates dense attention modules to refine features hierarchically, boosting detection performance across varied scenes.


Recently, FATCNet~\cite{TAES2024FATCNet} employs a dual-branch CNN-Transformer with adaptive fusion for effective local-global feature integration. Similarly, ILNet~\cite{TAES2025ILNet} enhances detection by emphasizing low-level salient features via lightweight fusion and dynamic aggregation. Moreover, MNHU~\cite{TAES2025MNHUNet} introduces a high-order UNet with nested multi-scale fusion to capture long-range dependencies and hierarchical representations. In addition, SFCANet~\cite{TAES2025SFCANet} combines spatial-frequency attention with deep supervised multi-task learning to improve background suppression and target discrimination.



While effective to a degree, \textbf{these methods treat all backgrounds as a single class}, \textit{ignoring the significant impact of different background semantics} on distinguishing between true targets and false alarms with similar appearances.
This simplification limits the ability of the model to exploit contextual cues embedded in semantically distinct regions such as sky, sea, or land, which often exhibit different noise characteristics and target distributions.

Moreover, although deep learning-based methods no longer explicitly rely on sparse priors or local contrast assumptions as traditional approaches do, many still implicitly incorporate such prior knowledge to improve detection performance. However, in dense scenes characterized by multiple closely spaced targets, the assumption of local contrast between a target and its surrounding background becomes invalid. This shift in spatial distribution not only weakens the effectiveness of contrast-based cues but also introduces significant ambiguity in distinguishing adjacent targets. Consequently, there is a pressing need for methods capable of modeling the complex interactions and contextual dependencies among targets in densely populated scenarios.

To address these limitations, our work explores a novel detection paradigm for infrared small targets, focusing on two main aspects:
\begin{enumerate}
    \item \textbf{Unexplored Challenge for Clustered Targets:} 
    Departing from conventional assumptions of sparsity, we target the more realistic and challenging scenario of detecting densely clustered targets. To support this direction, we construct DenseSIRST, a specialized dataset designed to benchmark performance under such conditions.
    \item \textbf{Explicit Semantic Modeling:} 
    Rather than treating all backgrounds uniformly, we utilize the semantic annotations provided by DenseSIRST to enable fine-grained background-aware detection. Our proposed BAFE-Net leverages this information to enhance contextual understanding, thereby improving the discrimination between targets and background distractors.
     
\end{enumerate}

\section{The DenseSIRST Dataset} \label{sec:dataset}



\begin{figure*}[hbtp]
\vspace{-1\baselineskip}
  \centering
  \includegraphics[width=0.98\linewidth]{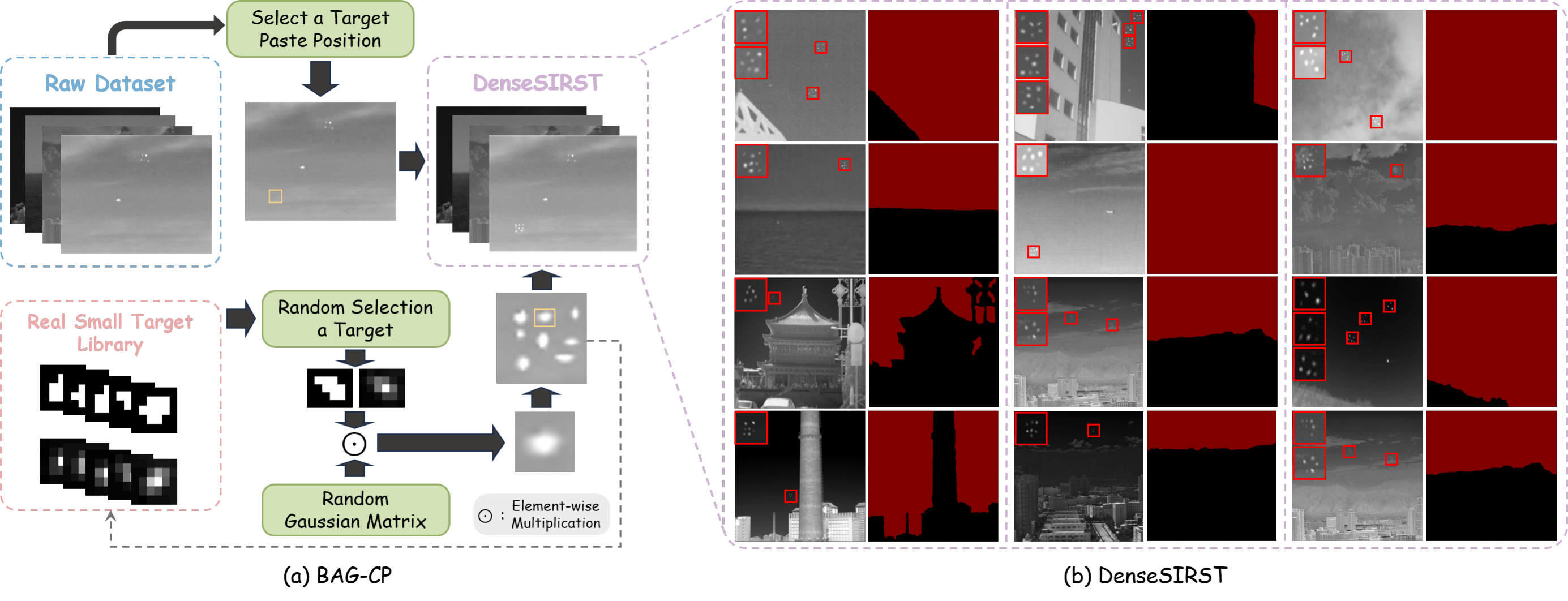}
  \vspace{-1\baselineskip}
  \caption{
  \textbf{Synthetic generation pipeline and representative samples of the DenseSIRST dataset.} (a) The BAG-CP pipeline for synthesizing the DenseSIRST dataset incorporates three key components: semantic-aware background selection, Gaussian-based target fusion, and realistic sample generation. (b) Illustration of selected images from our proposed DenseSIRST dataset. Each image pair consists of a simulated dense small target image (left) and its corresponding sky-segmented version (right).
  }
  \label{fig:DenseSIRST}
\vspace{-1\baselineskip}
\end{figure*}

Infrared small target detection is crucial in surveillance, early warning, and remote sensing. However, detecting small, low-contrast targets against complex and cluttered backgrounds remains a significant challenge. Existing benchmark datasets, such as SIRST V1~\cite{dai2021asymmetric}, SIRST V2~\cite{dai2023one}, IRSTD1K~\cite{zhang2022isnet}, and SIRSTAUG~\cite{zhang2023attention}, have contributed substantially to algorithmic advancement, yet they predominantly feature sparsely distributed targets and lack comprehensive pixel-level background annotations, as demonstrated in Table~\ref{tab:datasets}.

The mismatch between current benchmarks and real-world scenarios hinders the development of methods for detecting dense, low-contrast targets within complex and semantically rich backgrounds. To address this gap, we present DenseSIRST, a novel infrared dataset tailored for dense infrared small target detection under complex backgrounds. DenseSIRST offers high-density target distributions and fine-grained pixel-level annotations, as illustrated in Fig.~\ref{fig:DenseSIRST}~(b). By providing richer spatial and semantic context, it aims to foster the development of more robust and practically applicable detection algorithms.

\subsection{Dataset Construction}




The DenseSIRST dataset is constructed by integrating synthetically generated small targets into infrared background images from the publicly available SIRST V2~\cite{dai2023one} dataset. To achieve realistic and contextually consistent fusion, we propose a novel strategy called Background-Aware Gaussian Copy-Paste (BAG-CP), which builds upon the principles of mixup \cite{ICLR2018mixup}. Unlike naive copy paste, BAG-CP selectively pastes targets into semantically regions informed by background context, thereby enhancing the naturalness and variability of synthesized samples. The overall framework is depicted in Fig.~\ref{fig:DenseSIRST}~(a).

\subsubsection{Infrared Small Target Library Construction}
To simulate diverse and realistic small targets, we first build a target library by extracting individual small target patches from infrared datasets. Each extracted target is carefully cropped to include the target region with minimal background. This target library provides a rich source of varied target appearances, shapes, and sizes for subsequent synthesis.

\subsubsection{Semantic-Aware Target Paste Region Selection} 
To ensure plausible and contextually consistent target placement, the BAG-CP strategy incorporates semantic-aware selection of paste regions. Specifically, from each background image from SIRST V2, one to three candidate regions are randomly defined within relatively smooth and homogeneous sky areas. Each selected region is fixed to a size of $20 \times 20$ pixels and serves as a dense paste area for simulating clustered target distributions.

\subsubsection{Target Pasting and Fusion Process} 
Within each selected region, 6 to 12 targets are randomly sampled from the target library and pasted with controlled scale and spacing. Each target is resized to a dimension no larger than $5 \times 5$ pixels to mimic realistic scale variation. Targets are spaced with 1 to 2 pixels apart to emulate natural cluster formations typical in dense small target scenarios.
To seamlessly blend the pasted targets with the background, an adaptive Gaussian fusion matrix $G$ is applied, producing smooth transitions at target edges and avoiding harsh visual boundaries. The synthesized target patch $T_{new}$ is computed as:
\begin{align}
    T_{new} = B_{raw} + T_{raw} \times \lambda \times G  \ , \lambda \in [0.5, 1] , \label{eq:equation1}
\end{align}
where $B_{raw}$ is the original background patch, $T_{raw}$ is the raw target instance, and $\lambda$ is a random brightness scaling factor introducing intensity diversity. The Gaussian matrix $G$ is defined by
\begin{align}
G(x, y) = e^{-\frac{{(x - \rho_x w)^2}}{{2\sigma_x^2}} - \frac{{(y - \rho_y h)^2}}{{2\sigma_y^2}}} , \label{eq:equation2}
\end{align}
where $w$ and $h$ represent the width and height of the target, respectively. The parameters $\rho_x$ and $\rho_y$ specify the Gaussian center offset relative to the target dimensions, randomly sampled from $[0, 0.2]$. The standard deviations $\sigma_x$ and $\sigma_y$, controlling the spread of the Gaussian kernel, are independently sampled from $[0.3, 0.6]$, which enables varied edge softness and thus improves the naturalness of target boundaries.




\subsubsection{Geometric and Appearance Augmentation} 
To further increase target diversity, each pasted target undergoes a random rotation transformation with angle $\theta$ uniformly sampled from $[-90^\circ, 90^\circ]$, implemented by
\begin{align}
    \begin{bmatrix} x' \\ y' \end{bmatrix} = \begin{bmatrix} \cos(\theta) & -\sin(\theta) \\ \sin(\theta) & \cos(\theta) \end{bmatrix} \begin{bmatrix} x - \mu_x \\ y - \mu_y \end{bmatrix} ,  \label{eq:equation3}
\end{align}
where $(\mu_x, \mu_y)$ is the center of the target patch.

\begin{figure*}[htbp]
  \vspace{-1.5\baselineskip}
  \centering
  \subfloat[]{\includegraphics[height=0.241\textwidth]{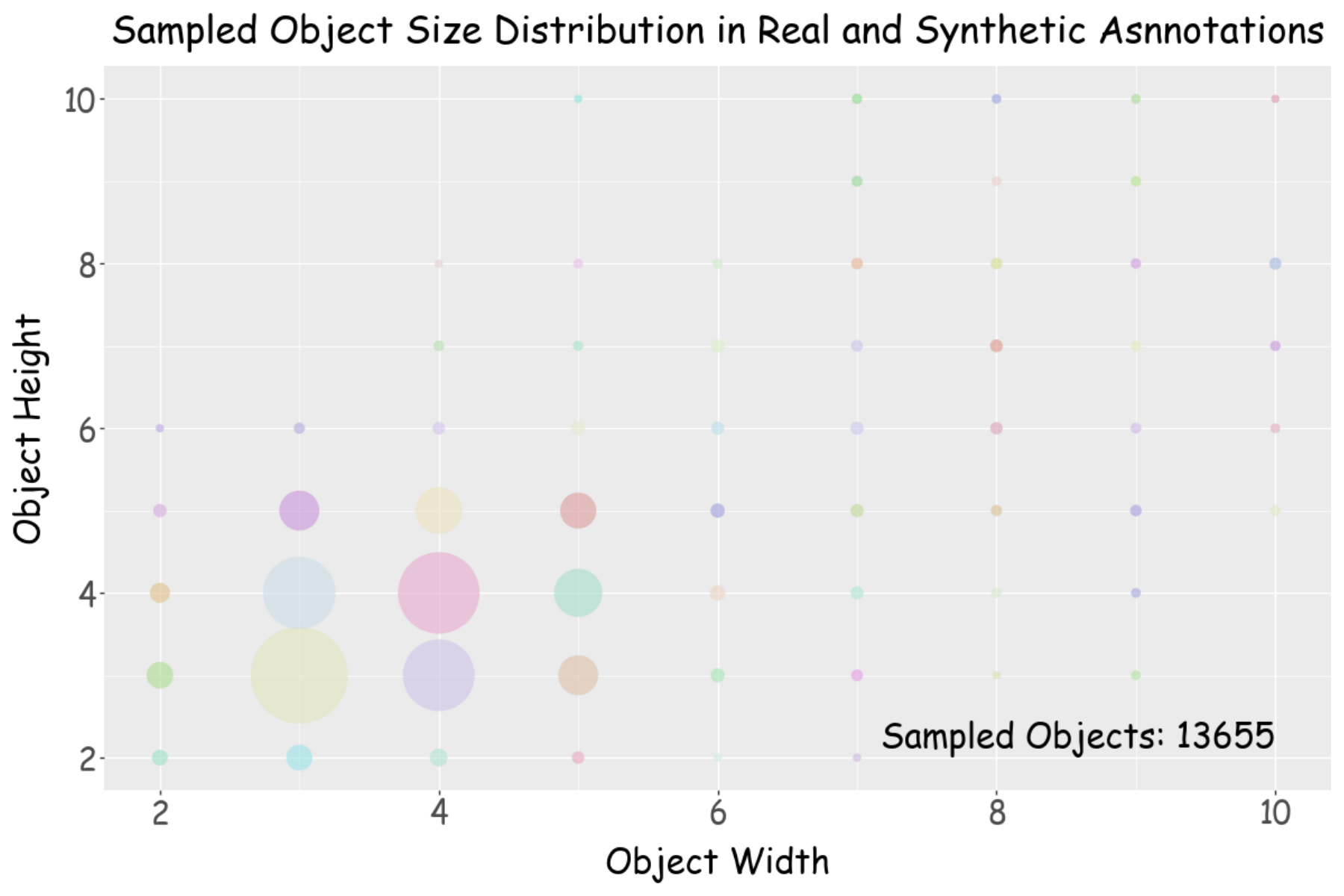}
  }
  \subfloat[]{\includegraphics[height=0.241\textwidth]{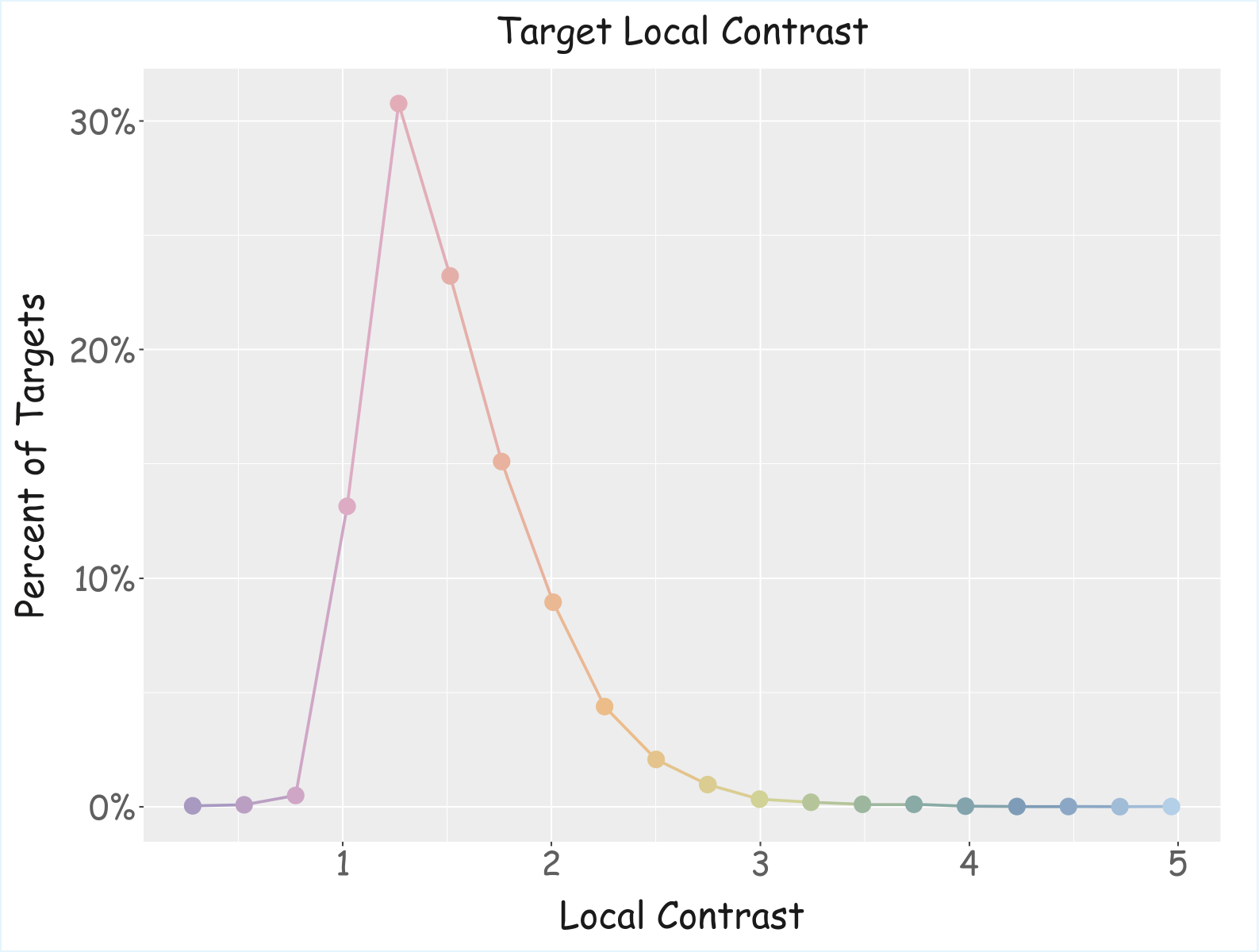}
  }
  \subfloat[]{\includegraphics[height=0.241\textwidth]{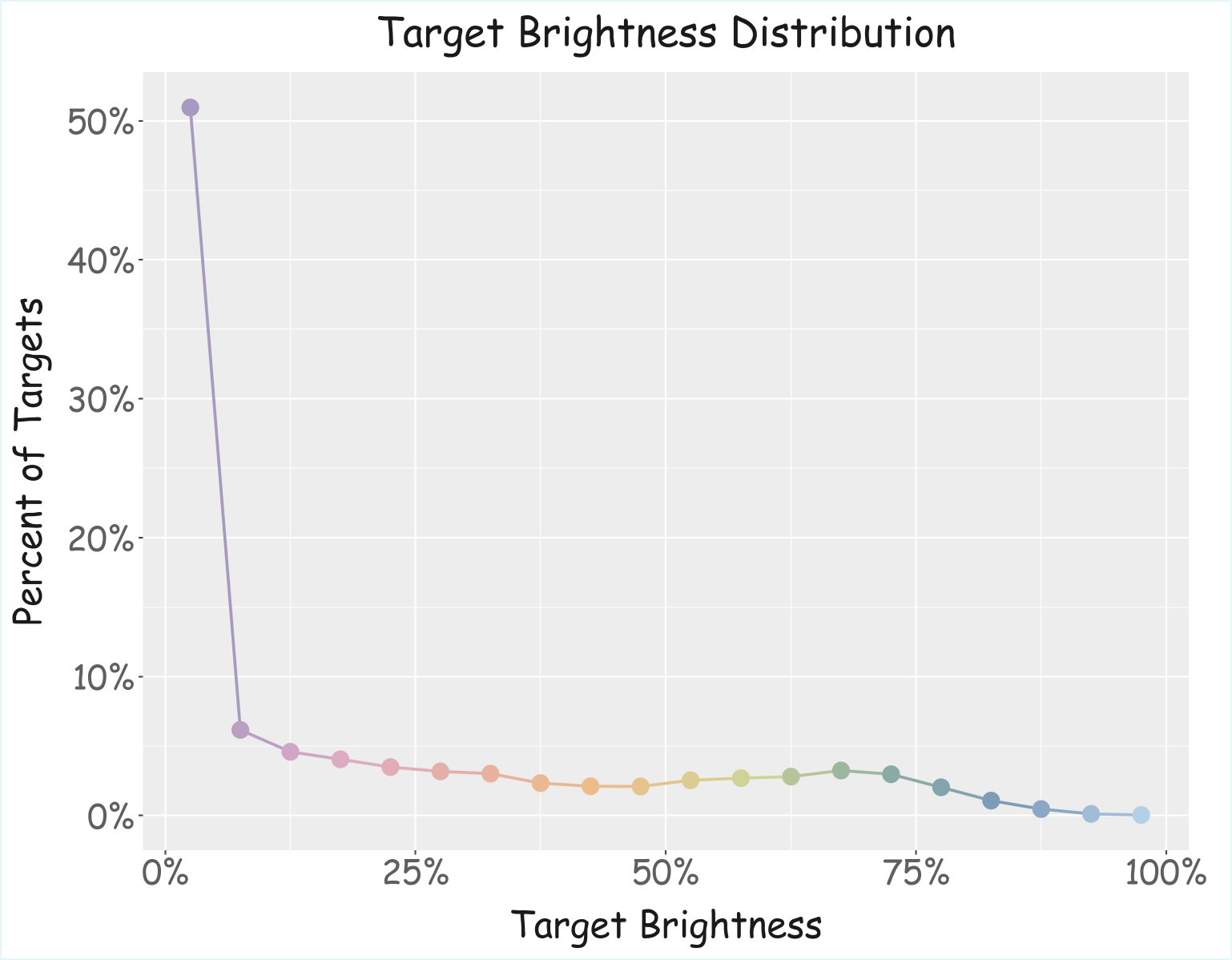}
  }
  \caption{
  \textbf{Statistical characteristics of our DenseSIRST dataset, underscoring the detection challenges it presents.} (a) The distribution of target sizes, with the circle sizes in the scatter plot indicating the prevalence of each size category. (b) The distribution of local contrasts for small targets in the dataset, demonstrating that the dataset encompasses a broad range of contrasts, down to a minimum of 0.15. (c) The brightness distribution of small targets in the dataset, highlighting the wide spectrum from dark to light targets this dataset offers for detection tasks.
  }
  \label{fig:distribution}
\end{figure*}

\subsection{Dataset Description and Characteristics}

\subsubsection{Dataset Composition}
The DenseSIRST dataset comprises 1024 infrared images, with a total of 13,655 densely clustered small targets. 
The images in the DenseSIRST dataset cover a wide range of realistic scenarios, including urban areas, mountainous regions, maritime environments, and cloudy scenes. This diversity enables the development and assessment of detection algorithms that can generalize well to various real-world applications. 
The division of training, validation, and test set is aligned with the SIRST V2 dataset. 

\subsubsection{Annotations and Labels}
In the DenseSIRST dataset, each image is meticulously annotated with three types of labels: pixel-level masks, bounding boxes, and point annotations. The pixel-level masks provide precise segmentation of each target, enabling the evaluation of algorithms at a fine-grained level. Bounding box annotations encapsulate each target with a rectangular box, facilitating the assessment of detection performance using metrics such as precision, recall, and F1 score. Point annotations mark the center of each target, allowing for the evaluation of localization accuracy.

\subsubsection{Density Definition}

To quantitatively capture the degree of target clustering, we define a density measure tailored for cluster regions within the dataset. A cluster region is identified when the Euclidean distance between the centroids of any two targets is less than 10 pixels, representing a meaningful spatial proximity threshold for dense grouping.

Formally, the density $\mathcal{D}$ of a cluster region is computed as:
\begin{align}
    \mathcal{D} = \mathcal{N} \times \frac{\mathcal{A}_{t}}{\mathcal{A}_{c}}, 
    \label{eq:equation4}
\end{align}
where $\mathcal{N}$ denotes the number of targets within the cluster, $\mathcal{A}{t}$ is the cumulative pixel area covered by all target masks inside the cluster, and $\mathcal{A}{c}$ represents the total pixel area of the cluster region itself. This metric jointly reflects both the quantity of targets and their relative spatial occupancy, providing a nuanced descriptor of local target density.

For consistent comparison across images, density values are normalized via min-max scaling based on the global minimum and maximum densities observed in both clustered and isolated target regions. Empirical statistics reveal a pronounced density difference, cluster regions exhibit an average density of 0.4092 compared to 0.1342 in single target regions, underscoring the metrics effectiveness in distinguishing densely grouped targets from sparse ones.



\subsubsection{Statistical Characteristics}

Fig.~\ref{fig:distribution} illustrates key statistical properties of the DenseSIRST dataset, emphasizing the detection challenges it entails.

\begin{itemize}

\item \textbf{Target Size Distribution (Fig.~\ref{fig:distribution}~(a)):} The vast majority of targets are extremely small, predominantly under $5 \times 5$ pixels, with the most frequent size being $ 3 \times 3$ pixels. This distribution aligns with the typical scale of small infrared targets.

\item \textbf{Local Contrast Distribution (Fig.~\ref{fig:distribution}~(b)):} Approximately 90\% of targets possess a local contrast below 2 relative to their surrounding background. This low contrast is indicative of the inherent difficulty in detecting such targets against cluttered backgrounds. Moreover, the presence of nearby targets within dense clusters can inadvertently be included in the background calculation, leading to an underestimation of local contrast and further complicating detection.

\item \textbf{Target Brightness Distribution (Fig.~\ref{fig:distribution}~(c)):} The brightness values of targets span a wide spectrum, with only a minority being the brightest points in their respective images. This diversity invalidates brightness as a reliable standalone feature, highlighting the necessity for robust algorithms that leverage richer contextual and structural information.

\end{itemize}

In conclusion, the DenseSIRST dataset exhibits rich annotation modalities, diverse environmental scenarios, and statistically challenging target characteristics. These attributes collectively establish it as a comprehensive and realistic benchmark, particularly well suited for developing and evaluating infrared small target detection algorithms under dense and low-contrast conditions.





\section{Background-Aware Feature Exchange Network} \label{sec:method}

\subsection{Overall Architecture and Training Pipeline}

\begin{figure*}[hbtp]
\vspace{-0.5\baselineskip}
  \centering
  \includegraphics[width=0.98\linewidth]{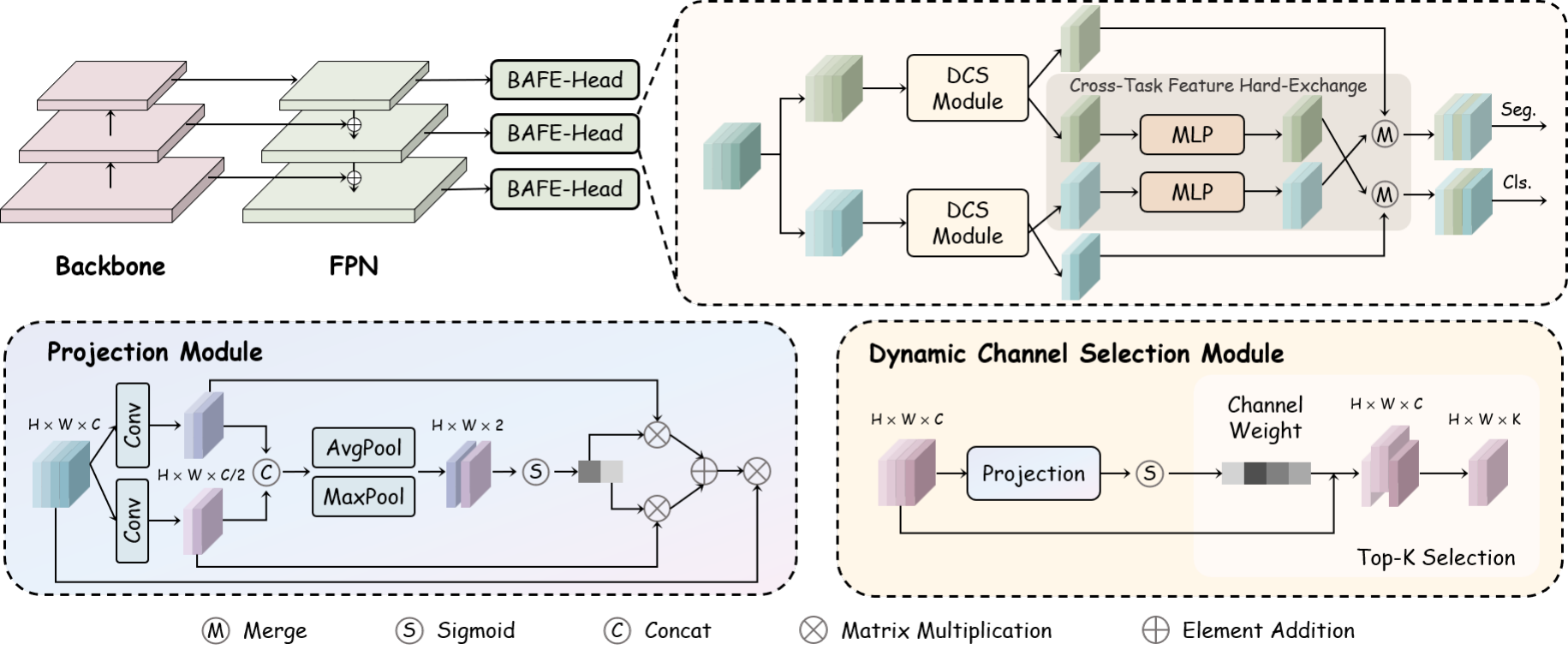}
  \vspace{-0.5\baselineskip}
  \caption{\textbf{The framework of BAFE-Net.} BAFE-Net contains three modules: ResNet, FPN and BAFE-Head. The Head module integrates a background segmentation branch to work in parallel with the classification branch. Both branches utilize the DCS Module to select the most discriminative top-k channel features, followed by channel-wise feature interaction. This enhancement transforms the original single-task object detection framework into a multi-task learning architecture, enabling simultaneous and efficient execution of both object detection and background semantic segmentation.
  }
  \label{fig:bafe-net}
\vspace{-1\baselineskip}
\end{figure*}

The proposed BAFE-Net architecture, which builds upon the fully convolutional one-stage object detector (FCOS) \cite{ICCV2019FCOS}, is illustrated in Fig.~\ref{fig:bafe-net}. The network comprises three key components: ResNet backbone, Feature Pyramid Network (FPN), and BAFE-Head module. The BAFE-Head features two parallel branches: the original FCOS detection branch and an innovative background segmentation branch. This dual-branch design enables explicit modeling of background semantics, which is essential for distinguishing real targets from background clutter in infrared scenes.

BAFE-Net introduces a dynamic cross-task feature hard-exchange mechanism that facilitates bidirectional information flow between the detection and segmentation branches. Both branches incorporate a Dynamic Channel Selection (DCS) module, which identifies and utilizes the top-k most discriminative channel features to enable efficient Channel-wise Feature Exchange. This dynamic exchange allows the network to adaptively leverage the most relevant features from each task, leading to improved detection performance.

To address the challenge of limited target samples, BAFE-Net adopts the dataset synthesis method BAG-CP as data augmentation strategy. By leveraging semantic context, this method pastes synthetic infrared small targets exclusively onto sky regions in real-scene backgrounds, with Gaussian smoothing applied to the edges for seamless integration. This context-aware synthesis and integration approach enables BAFE-Net to effectively recognize and handle these challenging conditions, ultimately setting a new standard for performance in infrared small target detection.

\subsection{Dynamic Cross-Task Feature Exchange Head}

Infrared small target detection presents inherent challenges due to the weak discriminative characteristics of targets and the prevalence of complex background clutter. Conventional single-branch detection frameworks often struggle to effectively distinguish true targets from background distractors, particularly in scenarios characterized by limited target information and ambiguous contextual cues.

To address these challenges, we propose a multi-task collaborative framework that integrates object detection and background semantic segmentation within a unified prediction head, referred to as BAFE-Head. The core motivation behind this design lies in the complementary nature of the two tasks: object detection is tailored to localize small, sparse targets, whereas semantic segmentation captures global scene semantics and background structures. By jointly optimizing these tasks, the proposed framework leverages background semantics to suppress false positives and utilizes target localization cues to refine segmentation boundaries. This joint formulation facilitates a more holistic and robust feature representation, enhancing the discriminative capability of the network in cluttered infrared scenes.

\subsubsection{Dynamic Channel Selection Module}

Unlike conventional multi-task learning strategies that typically employ static or full feature sharing, which may lead to the propagation of redundant or even conflicting information across tasks, we introduce a dynamic cross-task feature exchange mechanism. This mechanism adaptively selects and exchanges task-relevant channel-wise features between the detection and segmentation branches. The importance of each channel is estimated using an attention mechanism, which assigns weights to channels according to their contribution. This dynamic and selective interaction not only strengthens the synergy between detection and segmentation but also effectively mitigates the risk of negative transfer, thereby enhancing overall detection performance in complex infrared environments.

\textbf{Projection Module.} Given an input feature map $\mathbf{X} \in \mathbb{R}^{C \times H \times W}$, where $C$, $H$, and $W$ denote the number of channels, height, and width of the feature map, respectively, we design a Projection Module to generate a channel-wise attention vector $\mathbf{m} \in \mathbb{R}^{C}$. This design is inspired by our previous work, Large Selective Kernel Network (LSKNet) \cite{IJCV2024LSKNet}, and aims to enhance channel discrimination by integrating local feature extraction with global context modeling.

This is achieved by applying a series of operations to $\mathbf{X}$, including local feature extraction, channel-wise statistics computation, and weighted fusion.

\begin{itemize}

\item local feature extraction: To capture fine-grained local representations, we first apply two independent convolutional operations to the input feature map $\mathbf{X} \in \mathbb{R}^{C \times H \times W}$, each reducing the channel dimension to $C/2$. These two branches are designed to extract complementary local patterns.

\vspace{-0.5\baselineskip}
\begin{equation}
 \begin{aligned}
    \mathbf{X}_1 &= \text{Conv}_1(\mathbf{X}), \\
    \mathbf{X}_2 &= \text{Conv}_2(\mathbf{X}), \\
    \mathbf{X}_c &= \mathbf{X}_1 \oplus \mathbf{X}_2,
    \label{eq:equation5_1}
 \end{aligned}
\end{equation}

where $\oplus$ denotes concatenation operation, resulting in $\mathbf{X}_c \in \mathbb{R}^{C \times H \times W}$.

\item channel-wise statistics computation: To integrate global context and emphasize informative regions, we apply both average pooling and max pooling along the channel dimension to form a compact spatial descriptor, followed by a convolutional projection and sigmoid activation.

\begin{equation}
 \begin{aligned}
    \text{Feat} = \sigma(\mathcal{S}(\text{avg}(\mathbf{X}_c) \oplus \max(\mathbf{X}_c))),
    \label{eq:equation5_2}
 \end{aligned}
\end{equation}

where $\max$ and $\text{avg}$ are the maximum and average pooled spatial feature descriptors. $\mathcal{S}$ is a convolution operation that reduces the dimensionality of concatenated features, and $\sigma(\cdot)$ is the sigmoid function. The obtained $\text{Feat} \in \mathbb{R}^{2 \times H \times W}$ is a tensor, which represents the weighted fusion of $\max(\mathbf{X})$ and $\text{avg}(\mathbf{X})$.

\item weighted fusion: The attention map $\text{Feat}$ is split along the channel axis and used to reweight the previously extracted features $\mathbf{X}_1$ and $\mathbf{X}_2$ through element-wise multiplication. The reweighted features are then concatenated to form the output of the projection module:

\begin{equation}
 \begin{aligned}
    \mathbf{Z} &= \mathcal{F}(\mathbf{X}_1, \text{Feat}) \oplus \mathcal{F}(\mathbf{X}_2, \text{Feat}), \\
    \mathbf{m} &= \text{mean}\left(\sigma\left( \mathbf{X} \odot \mathbf{Z} \right)\right),
    \label{eq:equation5_3}
 \end{aligned}
\end{equation}

where $\odot$ denotes element-wise multiplication, and the operator $\mathcal{F}(\mathbf{X}_i, \text{Feat})$ denotes a channel-wise spatial reweighting operation that modulates the local feature $\mathbf{X}_i$ using the corresponding attention map $\text{Feat}_i$.
$\text{mean}$ denotes a global average pooling operation that aggregates feature representations across spatial dimensions, reducing spatial information into a channel-wise global descriptor. $\mathbf{m} \in \mathbb{R}^{C}$ is a one-dimensional tensor, and $C$ is the number of channels. Each element $\mathbf{m}$ represents the attention logits for the $i$-th channel.

\end{itemize}



For each sample, the top $K$ channels with the highest weights are selected, where $K$ is determined as a fraction $p$ of the total number of channels $C$:
\begin{equation}
\begin{aligned}
    K & = C \times p ,  
    \\
    {V}_{\text{topk}}, {I}_{\text{topk}} & = \mathrm{TopK}({\mathbf{m}}, K) ,  \label{eq:equation6}
\end{aligned}
\end{equation}
where ${V}_{\text{topk}} \in \mathbb{R}^{K}$ represents the top $K$ largest weight values for each sample. ${I}_{\text{topk}} \in \mathbb{N}^{K}$ represents the corresponding channel indices for these weights.
This dynamic selection mechanism enables adaptive identification of the most informative channels for each sample, thereby enhancing the effectiveness of cross-task feature interaction.

\begin{table*}[h]
  \renewcommand\arraystretch{1.2}
  \footnotesize
  \centering
    \vspace{-1\baselineskip}
  \caption{Comparison with Other State-of-the-art object detection methods on DenseSIRST.}
  \label{tab:sota}
  \setlength{\tabcolsep}{3pt}
  \begin{tabular}{l|c|c|c|cccc}
    \multirow{2}{*}{Method}  & \multirow{2}{*}{Backbone} & \multirow{2}{*}{FLOPs} & \multirow{2}{*}{Params} & \multicolumn{4}{c}{DenseSIRST} \\
      & & & & mAP\textsubscript{07} $\uparrow$ &  recall\textsubscript{07}  $\uparrow$ & mAP\textsubscript{12} $\uparrow$ &  recall\textsubscript{12} $\uparrow$  \\
  \Xhline{1pt}
  \multicolumn{5}{l}{\textit{End2End}}  \\
  \hline
    DETR  \cite{carion2020end} & ResNet50 & 24.940G & 41.555M & 0.000 & 0.000 & 0.000 & 0.000 \\
    Sparse R-CNN  \cite{sun2021sparse} & ResNet50 & 45.274G & 0.106G & 0.183 & 0.572 & 0.154 & 0.614\\
    Deformable DETR  \cite{zhu2020deformable} & ResNet50 & 51.772G & 40.099M & 0.024 & 0.016 & 0.018 & 0.197 \\
    Conditional DETR  \cite{meng2021conditional} & ResNet50 & 27.143G & 40.297M & 0.000 & 0.000 & 0.000 & 0.001\\
    DAB-DETR  \cite{liu2022dab} & ResNet50 & 28.939G & 43.702M & 0.005 & 0.054 & 0.000 & 0.001\\
  \hline
  \multicolumn{5}{l}{\textit{Two-stage}}  \\
  \hline
    Faster R-CNN  \cite{ren2015faster} & ResNet50 & 0.759T & 33.035M & 0.091 & 0.022 & 0.015 & 0.029 \\
    Cascade R-CNN  \cite{cai2019cascade} & ResNet50 & 90.978G & 69.152M & 0.136 & 0.188 & 0.139 & 0.194 \\
    Grid R-CNN  \cite{lu2019grid} & ResNet50 & 0.177T & 64.467M & 0.156 & 0.122 & 0.104 & 0.190 \\
    Libra R-CNN  \cite{pang2019libra} & ResNet50 & 63.990G & 41.611M & 0.141 & 0.142 & 0.085 & 0.120 \\
    TridentNet  \cite{li2019scale} & ResNet50 & 0.759T & 33.035M & 0.091 & 0.009 & 0.014 & 0.021 \\
    SABL  \cite{wang2020side} & ResNet50 & 0.125T & 42.213M & 0.124 & 0.104 & 0.104 & 0.171 \\
    Dynamic R-CNN  \cite{zhang2020dynamic} & ResNet50 & 63.179G & 41.348M & 0.184 & 0.235 & 0.111 & 0.190 \\
  \hline
  \multicolumn{5}{l}{\textit{One-stage}}  \\
  \hline
    SSD  \cite{liu2016ssd} & VGG16 & 87.552G & 23.746M & 0.211 & 0.421 & 0.178 & 0.424 \\
    RetinaNet  \cite{lin2017focal} & ResNet50 & 52.203G & 36.330M & 0.114 & 0.510 & 0.086 & 0.523 \\
    YOLOv3  \cite{farhadi2018yolov3} & Darknet & 50.002G & 61.949M & 0.233 & 0.424 & 0.207 & 0.413 \\
    CenterNet  \cite{zhou2019objects} & ResNet50 & 50.278G & 32.111M & 0.138 & 0.316 & 0.124 & 0.317 \\
    FCOS  \cite{ICCV2019FCOS} & ResNet50 & 50.291G & 32.113M & 0.232 & 0.315 & 0.204 & 0.324 \\
    ATSS  \cite{zhang2020bridging} & ResNet50 & 51.504G & 32.113M & 0.248 & 0.327 & 0.202 & 0.326 \\
    CentripetalNet  \cite{dong2020centripetalnet} & HourglassNet & 0.491T & 0.206G & 0.244 & 0.259 & 0.201 & 0.244 \\
    AutoAssign  \cite{zhu2020autoassign} & ResNet50 & 50.555G & 36.244M & 0.255 & 0.354 & 0.180 & 0.314 \\
    GFL  \cite{li2020generalized} & ResNet50 & 52.296G & 32.258M & \underline{0.264} & 0.367 & 0.230 & 0.317 \\
    PAA  \cite{kim2020probabilistic} & ResNet50 & 51.504G & 32.113M & 0.255 & 0.545 & 0.228 & 0.551 \\
    VFNet  \cite{zhang2021varifocalnet} & ResNet50 & 48.317G & 32.709M & 0.253 & 0.336 & 0.214 & 0.336 \\
    PVT-T  \cite{wang2021pyramid} & PVT & 41.623G & 21.325M & 0.109 & 0.481 & 0.093 & 0.501 \\
    YOLOF  \cite{chen2021you} & ResNet50 & 25.076G & 42.339M & 0.091 & 0.009 & 0.002 & 0.009 \\
    YOLOX  \cite{ge2021yolox} & CSPDarknet & 8.578G & 8.968M & 0.210 & 0.341 & 0.180 & 0.331 \\
    TOOD  \cite{feng2021tood} & ResNet50 & 50.456G & 32.018M & 0.256 & 0.355 & 0.226 & 0.342 \\
    DyHead  \cite{dai2021dynamic} & ResNet50 & 27.866G & 38.890M & 0.249 & 0.335 & 0.189 & 0.328 \\
    DDOD  \cite{chen2021disentangle} & ResNet50 & 46.514G & 32.378M & 0.253 & 0.335 & \underline{0.230} & 0.351 \\
    RTMDet  \cite{lyu2022rtmdet} & CSPNeXt & 51.278G & 52.316M & 0.229 & 0.349 & 0.212 & 0.350 \\
    EfficientDet  \cite{tan2020efficientdet} & EfficientNet & 34.686G & 18.320M & 0.146 & 0.464 & 0.094 & 0.517 \\
  \hline
    \rowcolor[rgb]{0.9,0.9,0.9}$\star$ \textbf{BAFE-Net (Ours)} & ResNet50 & 71.639G & 35.626M & \textbf{0.274} & 0.342 & \textbf{0.248} & 0.338 \\

\end{tabular}
\vspace{-0.5\baselineskip}
\end{table*}

\subsubsection{Cross-Task Feature Hard-Exchange}
The introduction of feature hard-exchange is fundamentally driven by the limitations of conventional feature fusion methods in infrared small target detection. Soft fusion techniques such as element-wise addition or concatenation often compromise discriminative power by diluting sparse target features through weighted averaging, while simultaneously permitting gradient conflicts between detection and segmentation objectives during backpropagation. Furthermore, the inherent semantic gap between heterogeneous tasks—object detection versus background segmentation—causes modality mismatch when transferring unadapted features directly. Feature hard-exchange directly addresses these challenges by enabling replacement of the most discriminative channels identified by DCS Module, ensuring explicit semantic transfer without feature dilution, maintaining gradient isolation through binary masking, and resolving cross-modal discrepancies via feature adaptation.

The cross-task feature hard-exchange mechanism operates through three sequential steps.

\begin{itemize}

\item Exchange Mask Generation: To perform feature exchange, let $\mathbf{X}_1$ and $\mathbf{X}_2$ be the input tensors from the target detection and background segmentation tasks, respectively, each with a shape of $(C, H, W)$, where $C$ is the number of channels, and $H$ and $W$ are the height and width. Using channel indices $I_{\text{topk}}$ from the DCS module (Eq.~\ref{eq:equation6}), we construct a binary exchange mask $\mathbf{e} \in \mathbb{R}^{C}$:

\begin{align}
    \mathbf{e}[i] = 
        \begin{cases}
        1, & \mathrm{if}\ i \in I_{\text{topk}} \\
        0,& \mathrm{otherwise}
        \end{cases} , \label{eq:equation7}
\end{align}

where $i$ denotes channel index. This mask $\mathbf{e}$ identifies feature channels for exchange. The 1D channel mask is broadcast to spatial dimensions:

\begin{align}
    \mathbf{E}_{d} = \mathbf{e} \otimes \mathbf{1}_{H \times W} ,  \label{eq:equation8}
\end{align}

where $\otimes$ indicates broadcasting operation, and $\mathbf{1}_{H \times W}$ is an all-one matrix. Detection and segmentation branches generate $\mathbf{E}_{d1}$ and $\mathbf{E}_{d2}$ respectively.

\item Feature Exchange and Adaptation: Bidirectional hard exchange is performed as:

\begin{align}
    \mathbf{Y}_1 = \mathbf{X}_1 \odot (\mathbf{1} - \mathbf{E}_{d1}) + \mathcal{A}\left(\mathbf{X}_2 \odot \mathbf{E}_{d2}\right) ,   \label{eq:equation9}\\
    \mathbf{Y}_2 = \mathbf{X}_2 \odot (\mathbf{1} - \mathbf{E}_{d2}) + \mathcal{A}\left(\mathbf{X}_1 \odot \mathbf{E}_{d1}\right) ,  \label{eq:equation10}
\end{align}

where $\odot$ denotes element-wise multiplication, $\mathbf{Y}_1$ and $\mathbf{Y}_2$ represent the output features, and $\mathbf{E}_{d1}$ and $\mathbf{E}_{d2}$ are the dynamic masks. To address the cross-modal nature of features, we introduce an adapter module $\mathcal{A}$, implemented as a multi-layer perceptron (MLP), which projects features into a common space before exchange. This design enables effective cross-modal interaction while preserving modality-specific characteristics.

\end{itemize}

Compared to soft fusion techniques, hard-exchange provides distinct advantages: complete channel replacement preserves feature integrity by avoiding partial feature loss; gradient isolation ensures that gradients propagate exclusively to the source task for exchanged channels; and binary masks enable computationally efficient zero-parameter exchange. This integrated design maintains high discriminative power in complex infrared scenarios while ensuring feature compatibility through the adapter $\mathcal{A}$.

\begin{table*}[h]
  \renewcommand\arraystretch{1.2}
  \footnotesize
  \centering
  \caption{Comparison with Other State-of-the-art object detection methods on IRSTD-1k and IRSDSS.}
  \label{tab:sota_other}
  \setlength{\tabcolsep}{3pt}
  \begin{tabular}{l|c|c|c|cccc|cccc}
    \multirow{2}{*}{Method}  & \multirow{2}{*}{Backbone} & \multirow{2}{*}{FLOPs} & \multirow{2}{*}{Params} & \multicolumn{4}{c}{IRSTD-1k} & \multicolumn{4}{c}{IRSDSS} \\
      & & & & mAP\textsubscript{07} $\uparrow$ &  recall\textsubscript{07}  $\uparrow$ & mAP\textsubscript{12} $\uparrow$ &  recall\textsubscript{12} $\uparrow$ & mAP\textsubscript{07} $\uparrow$ &  recall\textsubscript{07}  $\uparrow$ & mAP\textsubscript{12} $\uparrow$ &  recall\textsubscript{12} $\uparrow$  \\
  \Xhline{1pt}
  \multicolumn{5}{l}{\textit{End2End}}  \\
  \hline
    DETR  \cite{carion2020end} & ResNet50 & 24.940G & 41.555M & 0.000 & 0.000 & 0.000 & 0.000 & 0.000 & 0.001 & 0.000 & 0.001 \\
    Sparse R-CNN  \cite{sun2021sparse} & ResNet50 & 45.274G & 0.106G & 0.136 & 0.370 & 0.160 & 0.388 & 0.826 & 0.953 & 0.827 & 0.952 \\
    Deformable DETR  \cite{zhu2020deformable} & ResNet50 & 51.772G & 40.099M & 0.091 & 0.018 & 0.015 & 0.126 & 0.551 & 0.782 & 0.550 & 0.770 \\
    Conditional DETR  \cite{meng2021conditional} & ResNet50 & 27.143G & 40.297M & 0.000 & 0.001 & 0.000 & 0.002 & 0.000 & 0.010 & 0.000 & 0.011 \\
    DAB-DETR  \cite{liu2022dab} & ResNet50 & 28.939G & 43.702M & 0.002 & 0.002 & 0.000 & 0.005 & 0.346 & 0.769 & 0.219 & 0.762 \\
  \hline
  \multicolumn{5}{l}{\textit{Two-stage}}  \\
  \hline
    Faster R-CNN  \cite{ren2015faster} & ResNet50 & 0.759T & 33.035M & 0.298 & 0.346 & 0.269 & 0.344 & 0.416 & 0.401 & 0.478 & 0.524 \\
    Cascade R-CNN  \cite{cai2019cascade} & ResNet50 & 90.978G & 69.152M & 0.325 & 0.370 & 0.323 & 0.377 & 0.780 & 0.818 & 0.765 & 0.803 \\
    Grid R-CNN  \cite{lu2019grid} & ResNet50 & 0.177T & 64.467M & 0.299 & 0.402 & 0.306 & 0.393 & 0.746 & 0.826 & 0.720 & 0.786 \\
    Libra R-CNN  \cite{pang2019libra} & ResNet50 & 63.990G & 41.611M & 0.323 & 0.395 & 0.291 & 0.351 & 0.771 & 0.805 & 0.711 & 0.772 \\
    TridentNet  \cite{li2019scale} & ResNet50 & 0.759T & 33.035M & 0.311 & 0.341 & 0.288 & 0.352 & 0.672 & 0.757 & 0.631 & 0.701 \\
    SABL  \cite{wang2020side} & ResNet50 & 0.125T & 42.213M & 0.298 & 0.340 & 0.296 & 0.401 & 0.681 & 0.750 & 0.730 & 0.812 \\
    Dynamic R-CNN  \cite{zhang2020dynamic} & ResNet50 & 63.179G & 41.348M & 0.314 & 0.355 & 0.309 & 0.372 & 0.761 & 0.805 & 0.659 & 0.738 \\
  \hline
  \multicolumn{5}{l}{\textit{One-stage}}  \\
  \hline
    SSD  \cite{liu2016ssd} & VGG16 & 87.552G & 23.746M & 0.296 & 0.427 & 0.279 & 0.480 & 0.847 & 0.919 & 0.842 & 0.947 \\
    RetinaNet  \cite{lin2017focal} & ResNet50 & 52.203G & 36.330M & \underline{0.368} & 0.564 & 0.298 & 0.554 & 0.825 & 0.933 & 0.842 & 0.936 \\
    YOLOv3  \cite{farhadi2018yolov3} & Darknet & 50.002G & 61.949M & 0.200 & 0.311 & 0.158 & 0.342 & 0.779 & 0.918 & 0.830 & 0.881 \\
    CenterNet  \cite{zhou2019objects} & ResNet50 & 50.278G & 32.111M & 0.201 & 0.497 & 0.202 & 0.544 & 0.853 & 0.924 & 0.847 & 0.916 \\
    FCOS  \cite{ICCV2019FCOS} & ResNet50 & 50.291G & 32.113M & 0.307 & 0.426 & 0.303 & 0.446 & 0.841 & 0.907 & 0.842 & 0.924 \\
    ATSS  \cite{zhang2020bridging} & ResNet50 & 51.504G & 32.113M & 0.355 & 0.429 & 0.355 & 0.475 & 0.849 & 0.911 & 0.861 & 0.907 \\
    CentripetalNet  \cite{dong2020centripetalnet} & HourglassNet	 & 0.491T & 0.206G & 0.232 & 0.333 & 0.245 & 0.375 & 0.850 & 0.918 & 0.855 & 0.911 \\
    AutoAssign  \cite{zhu2020autoassign} & ResNet50 & 50.555G & 36.244M & 0.326 & 0.416 & 0.336 & 0.433 & 0.842 & 0.918 & 0.851 & 0.925 \\
    GFL  \cite{li2020generalized} & ResNet50 & 52.296G & 32.258M & 0.269 & 0.337 & 0.356 & 0.478 & 0.857 & 0.917 & 0.864 & 0.924 \\
    PAA  \cite{kim2020probabilistic} & ResNet50 & 51.504G & 32.113M & 0.356 & 0.668 & 0.342 & 0.573 & 0.957 & 0.941 & \underline{0.868} & 0.933 \\
    VFNet  \cite{zhang2021varifocalnet} & ResNet50 & 48.317G & 32.709M & 0.360 & 0.427 & \underline{0.362} & 0.456 & 0.848 & 0.908 & 0.854 & 0.912 \\
    PVT-T  \cite{wang2021pyramid} & PVT & 41.623G & 21.325M & 0.360 & 0.546 & 0.360 & 0.503 & 0.843 & 0.931 & 0.855 & 0.933 \\
    YOLOF  \cite{chen2021you} & ResNet50 & 25.076G & 42.339M & 0.216 & 0.304 & 0.146 & 0.276 & 0.341 & 0.441 & 0.281 & 0.429 \\
    YOLOX  \cite{ge2021yolox} & CSPDarknet & 8.578G & 8.968M & 0.367 & 0.533 & 0.316 & 0.476 & 0.850 & 0.917 & 0.844 & 0.911 \\
    TOOD  \cite{feng2021tood} & ResNet50 & 50.456G & 32.018M & 0.297 & 0.390 & 0.286 & 0.427 & \underline{0.860} & 0.914 & 0.864 & 0.908 \\
    DyHead  \cite{dai2021dynamic} & ResNet50 & 27.866G & 38.890M & 0.350 & 0.431 & 0.296 & 0.389 & 0.846 & 0.925 & 0.849 & 0.923 \\
    DDOD  \cite{chen2021disentangle} & ResNet50 & 46.514G & 32.378M & 0.281 & 0.362 & 0.338 & 0.456 & 0.855 & 0.921 & 0.856 & 0.928 \\
    RTMDet  \cite{lyu2022rtmdet} & CSPNeXt & 51.278G & 52.316M & 0.276 & 0.476 & 0.216 & 0.450 & 0.332 & 0.765 & 0.353 & 0.760 \\
    EfficientDet  \cite{tan2020efficientdet} & EfficientNet & 34.686G & 18.320M & 0.279 & 0.442 & 0.359 & 0.560 & 0.849 & 0.934 & 0.829 & 0.911 \\
  \hline
    \rowcolor[rgb]{0.9,0.9,0.9}$\star$ \textbf{BAFE-Net (Ours)} & ResNet50 & 71.639G & 35.626M & \textbf{0.410} & 0.545 & \textbf{0.370} & 0.497 & \textbf{0.860} & 0.900 & \textbf{0.868} & 0.938 \\

\end{tabular}
\vspace{-0.5\baselineskip}
\end{table*}

\begin{table*}[h]
  \renewcommand\arraystretch{1.2}
  \footnotesize
  \centering
    \vspace{-1\baselineskip}
  \caption{Comparison with Other State-of-the-art infrared target detection methods on DenseSIRST, IRSTD-1k and IRSDSS.}
  \label{tab:sota-infrared}
  \vspace{-0.5\baselineskip}
  \setlength{\tabcolsep}{3pt}
  \begin{tabular}{l|cccc|cccc|cccc}
    \multirow{2}{*}{Method}  & \multicolumn{4}{c}{DenseSIRST} & \multicolumn{4}{c}{IRSTD-1k} & \multicolumn{4}{c}{IRSDSS} \\
      & mAP\textsubscript{07} $\uparrow$ &  recall\textsubscript{07}  $\uparrow$ & mAP\textsubscript{12} $\uparrow$ &  recall\textsubscript{12} $\uparrow$ & mAP\textsubscript{07} $\uparrow$ &  recall\textsubscript{07}  $\uparrow$ & mAP\textsubscript{12} $\uparrow$ &  recall\textsubscript{12} $\uparrow$ & mAP\textsubscript{07} $\uparrow$ &  recall\textsubscript{07}  $\uparrow$ & mAP\textsubscript{12} $\uparrow$ &  recall\textsubscript{12} $\uparrow$  \\
  \Xhline{1pt}
    ACM       \cite{dai2021asymmetric} & 0.116 & 0.163 & 0.115 & 0.060 & 0.187 & 0.307 & 0.146 & 0.210 & 0.475 & 0.560 & 0.460 & 0.452 \\
    DNANet    \cite{li2022dense} & 0.188 & 0.219 & 0.135 & 0.139 & 0.161 & 0.245 & 0.196 & 0.220 & 0.705 & 0.807 & 0.786 & 0.789 \\
    AGPCNet   \cite{zhang2023attention} & 0.154 & 0.183 & 0.171 & 0.139 & 0.222 & 0.268 & 0.212 & 0.255 & 0.532 & 0.645 & 0.560 & 0.590 \\
    UIUNet    \cite{TIP2022UIUNet} & 0.178 & 0.235 & \underline{0.208} & 0.198 & \underline{0.349} & 0.475 & 0.192 & 0.175 & \underline{0.771} & 0.761 & \underline{0.831} & 0.826 \\
    MTUNet    \cite{TGRS2023MTUNet} & \underline{0.272} & 0.337 & 0.197 & 0.152 & 0.125 & 0.289 & \underline{0.242} & 0.344 & 0.627 & 0.737 & 0.709 & 0.707 \\
    RDIAN     \cite{TGRS2023RDIAN} & 0.101 & 0.124 & 0.172 & 0.128 & 0.107 & 0.149 & 0.115 & 0.138 & 0.650 & 0.754 & 0.694 & 0.731 \\
    URANet    \cite{wang2023URANet} & 0.201 & 0.224 & 0.185 & 0.155 & 0.181 & 0.570 & 0.200 & 0.571 & 0.694 & 0.794 & 0.807 & 0.832 \\
    ABCNet    \cite{ICME2023ABCNet} & 0.261 & 0.315 & 0.189 & 0.159 & 0.335 & 0.406 & 0.202 & 0.196 & 0.684 & 0.842 & 0.763 & 0.786 \\
    SeRankDet \cite{TGRS2024SeRankDet} & 0.216 & 0.261 & 0.114 & 0.120 & 0.308 & 0.408 & 0.160 & 0.268 & 0.658 & 0.742 & 0.774 & 0.788 \\
  \hline
    \rowcolor[rgb]{0.9,0.9,0.9}$\star$ \textbf{BAFE-Net (Ours)} & \textbf{0.274} & 0.342 & \textbf{0.248} & 0.338 & \textbf{0.410} & 0.545 & \textbf{0.370} & 0.497 & \textbf{0.860} & 0.900 & \textbf{0.868} & 0.938 \\

\end{tabular}
\vspace{-1.0\baselineskip}
\end{table*}

\section{Experiments} \label{sec:experiment}

\subsection{Experimental Settings} \label{subsec:setting}

\textit{1) \textbf{Dataset}:} During the experimental phase of our research, we conducted tests on the proposed dataset DenseSIRST. In addition to our proposed dataset, we also employed the publicly available IRSTD-1k dataset to further validate the effectiveness and generalization capability of our method.

\textit{2) \textbf{Implementation Details}:} 
Considering the image sizes in the dataset and computational overhead, for methods within the DeepInfrared toolkit, we set the range of the long edge of training and testing images to [512, 512] while maintaining aspect ratio to avoid image distortion. We choose Focal Loss as the loss function of the classification branch, IoU Loss as the loss function of the regression branch, and Cross Entropy Loss as the loss function of the segmentation branch, with equal weighting assigned to three losses to ensure balanced multi-task optimization.
We employ DAdaptAdam, a parameter-free adaptive optimization algorithm that dynamically adjusts learning rates based on gradient statistics, thereby improving training stability and reducing the need for manual learning rate tuning. The base learning rate is set to 1.0, and the weight decay is set to 0.05. To balance training time and convergence, we train all models for 24 epochs within the DeepInfrared framework.


\textit{3) \textbf{Evaluation Metrics}:}We used mean Average Precision (mAP) with IoU thresholds of 0.5 ($mAP_{07}$) and 0.75 ($mAP_{12}$), and Recall as evaluation metrics for our experiment. 

\begin{table}[htbp]
  \setlength{\abovecaptionskip}{0cm}  
  \renewcommand\arraystretch{1.2}
  \footnotesize
  \centering
  \vspace{-1.5\baselineskip}
  \caption{Ablation study on the impact of Segmentation Head, Feature Exchange and Adapter on detection performance in Background-Aware Feature Exchange}
  \label{tab:FeatureExchange-ablation}
  \setlength{\tabcolsep}{2pt}
  \begin{tabular}{c|ccc|cccc}
      \multirow{2}{*}{Strategy} & \multicolumn{3}{c|}{Module} & \multicolumn{4}{c}{DenseSIRST} \\
      & \makecell{Seg \\ Head} & \makecell{Feature \\ Exchange} & Adapter & mAP\textsubscript{07} $\uparrow$ &  recall\textsubscript{07}  $\uparrow$ & mAP\textsubscript{12} $\uparrow$ &  recall\textsubscript{12} $\uparrow$  \\
  \Xhline{1pt}
(a) & \xmark & \xmark & \xmark & 0.225  & 0.311 &  0.180 &  0.314 \\
  (b) & $\checkmark$ & \xmark & \xmark & 0.246  & 0.324 &  0.183 &  0.304 \\
  (c) & $\checkmark$ & $\checkmark$ & \xmark  & 0.254  & 0.321 & 0.199 & 0.310  \\
  \rowcolor[rgb]{0.9,0.9,0.9} (d) & $\checkmark$ & $\checkmark$ & $\checkmark$  & \textbf{0.262}  & 0.318 & \textbf{0.208} &  0.305 \\
  
\end{tabular}
\vspace{-1.0\baselineskip}
\end{table}

\begin{table}[htbp]
  \setlength{\abovecaptionskip}{0cm}  
  \renewcommand\arraystretch{1.2}
  \footnotesize
  \centering
  \vspace{-0.8\baselineskip}
  \caption{Ablation study on the impact of Hidden Layers on detection performance in Adapter}
  \label{tab:MLPHiddenLayer-ablation}
  \setlength{\tabcolsep}{4pt}
  \begin{tabular}{c|c|cccc}
      \multirow{2}{*}{Strategy} & \multirow{2}{*}{Hidden Layers} & \multicolumn{4}{c}{DenseSIRST} \\
      & & mAP\textsubscript{07} $\uparrow$ &  recall\textsubscript{07}  $\uparrow$ & mAP\textsubscript{12} $\uparrow$ &  recall\textsubscript{12} $\uparrow$  \\
  \Xhline{1pt}
  (a) & 8 & 0.243  & 0.327 &  0.206 &  0.308 \\
  (b) & 16 & 0.252  & 0.337 &  0.214 &  0.310 \\
  (c) & 32 & 0.254  & 0.326 & 0.218 & 0.312  \\
\rowcolor[rgb]{0.9,0.9,0.9} (d) & 64 & \textbf{0.260}  & 0.329 & \textbf{0.230} &  0.323 \\
  (e) & 128 & 0.254  & 0.326 &  0.211 &  0.333 \\
  (f) & 256 & 0.243  & 0.319 &  0.212 &  0.312 \\
  
\end{tabular}
\vspace{-1\baselineskip}
\end{table}

\begin{table}[htbp]
  \setlength{\abovecaptionskip}{0cm}  
  \renewcommand\arraystretch{1.2}
  \footnotesize
  \centering
  \vspace{-1\baselineskip}
  \caption{Ablation study on the impact of Exchange Method on detection performance in Feature Exchange Mechanism}
  \label{tab:ExchangeMechanism-ablation}
  \setlength{\tabcolsep}{3pt}
  \begin{tabular}{c|c|cccc}
      \multirow{2}{*}{Strategy} & \multirow{2}{*}{Exchange Method} & \multicolumn{4}{c}{DenseSIRST} \\
      & & mAP\textsubscript{07} $\uparrow$ &  recall\textsubscript{07}  $\uparrow$ & mAP\textsubscript{12} $\uparrow$ &  recall\textsubscript{12} $\uparrow$  \\
  \Xhline{1pt}
\rowcolor[rgb]{0.9,0.9,0.9} (a) & Channel & \textbf{0.266}  & 0.336 &  \textbf{0.230} &  0.323 \\
  (b) & Spatial & 0.223  & 0.304 &  0.218 &  0.322 \\
  (c) & Channel + Spatial & 0.246  & 0.310 & 0.199 & 0.314  \\
  (d) & Spatial + Channel & 0.224  & 0.310 & 0.201 &  0.328 \\
  
\end{tabular}
\vspace{-1\baselineskip}
\end{table}

\begin{table}[htbp]
  \setlength{\abovecaptionskip}{0cm}  
  \renewcommand\arraystretch{1.2}
  \footnotesize
  \centering
  \vspace{-0.5\baselineskip}
  \caption{Ablation study on the impact of feature selection on detection performance in Channel-wise Feature Exchange}
  \label{tab:DynamicExchange}
  \setlength{\tabcolsep}{3pt}
  \begin{tabular}{c|c|cccc}
      \multirow{2}{*}{Strategy} & \multirow{2}{*}{Feature Selection} & \multicolumn{4}{c}{DenseSIRST} \\
      & & mAP\textsubscript{07} $\uparrow$ &  recall\textsubscript{07}  $\uparrow$ & mAP\textsubscript{12} $\uparrow$ &  recall\textsubscript{12} $\uparrow$  \\
  \Xhline{1pt}
  (a) & Random Selection & 0.255 & 0.322 & 0.172 & 0.317 \\
  (b) & Fixed Selection & 0.273 & 0.333 & 0.230 & 0.323  \\
    \rowcolor[rgb]{0.9,0.9,0.9} (c) & Dynamic Selection & \textbf{0.283} & 0.335 & \textbf{0.233} & 0.325 \\
  
\end{tabular}
\vspace{-1\baselineskip}
\end{table}

\begin{table}[htbp]
  \setlength{\abovecaptionskip}{0cm}  
  \renewcommand\arraystretch{1.2}
  \footnotesize
  \centering
  \vspace{-0.5\baselineskip}
  \caption{Ablation study on the impact of the number of exchange channels on detection performance in Top-K selection}
  \label{tab:Top-K}
  \setlength{\tabcolsep}{4pt}
  \begin{tabular}{c|c|cccc}
      \multirow{2}{*}{Strategy} & \multirow{2}{*}{P} & \multicolumn{4}{c}{DenseSIRST} \\
      & & mAP\textsubscript{07} $\uparrow$ &  recall\textsubscript{07}  $\uparrow$ & mAP\textsubscript{12} $\uparrow$ &  recall\textsubscript{12} $\uparrow$  \\
  \Xhline{1pt}
  \rowcolor[rgb]{0.9,0.9,0.9} (a) & 1/2 & \textbf{0.283} & 0.335 & \textbf{0.233} & 0.325 \\
  (b) & 1/4 & 0.241 & 0.316 & 0.230 & 0.331 \\
  (c) & 3/4 & 0.242 & 0.325 & 0.204 & 0.310  \\
  (d) & 1/8 & 0.238 & 0.312 & 0.218 & 0.323 \\
  (e) & 3/8 & 0.255 & 0.327 & 0.203 & 0.331 \\
  (f) & 5/8 & 0.259 & 0.325 & 0.201 & 0.328  \\
  (g) & 7/8 & 0.238 & 0.316 & 0.214 & 0.329 \\
  
\end{tabular}
\vspace{-1\baselineskip}
\end{table}

\begin{table*}[htbp]
  \setlength{\abovecaptionskip}{0cm}  
  \renewcommand\arraystretch{1.2}
  \footnotesize
  \centering
  \vspace{-1\baselineskip}
  \caption{Ablation study on the impact of different ResNet on detection performance in BAFE-Net}
  \label{tab:Resnet}
  \setlength{\tabcolsep}{4pt}
  \begin{tabular}{c|c|cccc|cccc}
      \multirow{2}{*}{Strategy} & \multirow{2}{*}{ResNet} & \multicolumn{4}{c}{DenseSIRST} & \multicolumn{4}{c}{IRSTD-1k} \\
      & & mAP\textsubscript{07} $\uparrow$ &  recall\textsubscript{07}  $\uparrow$ & mAP\textsubscript{12} $\uparrow$ &  recall\textsubscript{12} $\uparrow$ & mAP\textsubscript{07} $\uparrow$ &  recall\textsubscript{07}  $\uparrow$ & mAP\textsubscript{12} $\uparrow$ &  recall\textsubscript{12} $\uparrow$ \\
  \Xhline{1pt}
  \rowcolor[rgb]{0.9,0.9,0.9} (a) & ResNet18 & \textbf{0.283} & 0.335 & 0.233 & 0.325 & \textbf{0.433} & 0.603 & \underline{0.380} & 0.528 \\
  (b) & ResNet34 & \underline{0.281} & 0.354 & 0.226 & 0.332 & 0.385 & 0.495 & \textbf{0.381} & 0.564 \\
  \rowcolor[rgb]{0.9,0.9,0.9} (c) & ResNet50 & 0.274 & 0.342 & \textbf{0.248} & 0.338 & \underline{0.410} & 0.545 & 0.370 & 0.497 \\
  (d) & ResNet101 & 0.261 & 0.317 & \underline{0.241} & 0.336 & 0.385 & 0.520 & 0.378 & 0.561 \\

\end{tabular}
\end{table*}

\subsection{Comparison with State-of-the-Arts} \label{subsec:sota}

We benchmarked our BAFE-Net against current leading or state-of-the-art methods on 
the proposed DenseSIRST dataset, 
using metrics such as $mAP_{07}$, $mAP_{12}$, and Recall. The experimental results 
for general object detection frameworks 
are presented in Table~\ref{tab:sota}.

End-to-end (End2End) methods, while theoretically appealing due to their NMS-free style, demonstrated suboptimal performance 
on both datasets. 
This can be attributed to their inherent requirement for extensive training data, akin to COCO-level volumes, for effective operation. The relatively limited availability of infrared small target detection datasets inherently handicaps such methods, resulting in their diminished performance.

Next, we observed that two-stage methods underperformed compared to one-stage methods. Two-stage methods generate candidate regions based on deeper features, which are then subjected to classification and bounding box regression. However, the DenseSIRST dataset, characterized by small and densely clustered infrared targets, poses a challenge for such methods, leading to their suboptimal performance.

In contrast, one-stage methods, which directly regress bounding boxes and categories on the feature pyramids, demonstrated superior performance 
on both datasets. 
These methods are not only faster and more computationally efficient, but also achieved higher mAP and recall, indicating their overall effectiveness.
Significantly, our BAFE-Net method outperformed all other one-stage methods 
on both datasets. 

Beyond the DenseSIRST dataset, we further conducted experiments on the IRSTD-1k\cite{zhang2022isnet} and IRSDSS\cite{TGRS2024irsdss} datasets, as summarized in Table~\ref{tab:sota_other}. The results on IRSTD-1k were consistent with those observed on DenseSIRST, again highlighting the advantage of BAFE-Net in handling dense small-target detection. In contrast, the IRSDSS dataset mainly consists of maritime targets with relatively large object sizes. Consequently, while BAFE-Net still ranked among the best-performing approaches, the differences among the top methods on this dataset were relatively minor. Notably, BAFE-Net achieved state-of-the-art performance, with mAP scores matching or surpassing those of the best competing methods.

The superior performance of BAFE-Net consistently validates the efficacy of our approach in accurately detecting small infrared targets under diverse detection scenarios, including both densely clustered and sparsely distributed configurations.


Furthermore, to comprehensively situate BAFE-Net within the specific domain of infrared small target detection (IRSTD), we conducted comparisons against state-of-the-art methods specifically designed for this task. The results of these comparisons on both DenseSIRST, IRSTD-1k and IRSDSS datasets are presented in Table~\ref{tab:sota-infrared}. As shown, BAFE-Net also achieves superior performance compared to these specialized IRSTD methods, further demonstrating its effectiveness.



\subsection{Visual Analysis} \label{subsec:visualization}

\begin{figure*}[hbtp]
  \centering
  \includegraphics[width=0.95\linewidth]{figs/experiment/vis-2.pdf}
  \vspace{-0.5\baselineskip}
  \caption{\textbf{Feature visualization comparison.} 
  This comparison showcases the detection results of different methods in terms of feature visualization. The proposed BAFE-Net demonstrates superior performance by effectively detecting small objects with fewer false alarms.
  }
  \label{fig:heatmap}
\vspace{-0.5\baselineskip}
\end{figure*}

\subsection{Ablation Study} \label{subsec:ablation}

\subsubsection{\textbf{Ablation Study on the Scheme of Background-Aware Feature Exchange}}


The BAFE module constitutes the core contribution of this paper, as evidenced by its significant impact on detection performance demonstrated in Table~\ref{tab:FeatureExchange-ablation}. To validate the effectiveness of the proposed components, we conducted an extensive ablation study by exploring various combinations of the background-based segmentation, Feature Exchange module, and adapter module.

Table~\ref{tab:FeatureExchange-ablation} summarizes the results and reveals several key insights. First, each individual module contributes significantly and independently to detection performance, thereby validating the foundational design of our framework. Second, a comparison between Strategy (a) and Strategy (b) highlights the benefit of incorporating background-based segmentation into the object detection pipeline, where the detection head interacts with the segmentation head. This integration results in performance gains of 2.1\% and 0.3\% on the DenseSIRST dataset.

Furthermore, the performance difference between Strategy (b) and Strategy (c) demonstrates the effectiveness of the Feature Exchange mechanism in enhancing the interaction between the detection and segmentation heads. This enhancement leads to improvements of 0.8\% and 1.6\% on the DenseSIRST dataset. Finally, comparing Strategy (c) with Strategy (d) shows that introducing a feature adapter prior to the Feature Exchange further amplifies detection performance.

As the feature adapter employs a Multi-Layer Perceptron (MLP), the number of hidden layers in the MLP is a crucial parameter that warrants careful consideration. Increasing the number of hidden layers typically improves performance but comes at the cost of increased computational complexity. To strike a balance between performance and efficiency, we conducted an experiment to determine the optimal number of hidden layers, as shown in Table~\ref{tab:MLPHiddenLayer-ablation}. 
The results show that performance consistently improves as the number of hidden layers increases, up to 64 layers. However, when the number of hidden layers exceeds 64, the detection performance begins to decline, while the parameter count continues to grow significantly. Therefore, we select 64 hidden layers for the Adapter, as it offers the best trade-off between detection accuracy and computational overhead.



\subsubsection{\textbf{Ablation Study on Feature Exchange Mechanism}}



In this section, we conduct a rigorous investigation of various Feature Exchange Mechanisms, including Channel Exchange, Spatial Exchange, and several of their combinations. The goal is to identify the most effective strategy for facilitating cross-task feature interaction. Channel Exchange enables direct information flow between features at the channel level, while Spatial Exchange operates by exchanging information along the spatial dimensions of the feature maps. To ensure an equitable exchange of features, the exchange ratio is consistently fixed at 0.5 in all configurations.

As shown in Table~\ref{tab:ExchangeMechanism-ablation}, Channel Exchange consistently outperforms other Feature Exchange strategies. The experimental results clearly highlight its effectiveness and stability, confirming its selection as the optimal mechanism for feature exchange within our model.

\subsubsection{\textbf{Ablation Study in Dynamic Channel Selection}}

In this section, we perform an in-depth investigation of different feature selection strategies, including random selection, fixed selection, and dynamic selection, with the goal of identifying the most effective approach for enhancing feature exchange. 
\textit{1) Randomly Selection}: This method selects feature channels randomly, which may lead to suboptimal feature exchanges. In some cases, it could even select features that hinder the task, ultimately reducing model performance. \textit{2) Fixed Selection}: While fixed selection provides consistency during training and eliminates uncertainty, it lacks the adaptability to dynamic input features, thus limiting the model's performance in complex or varied scenarios. \textit{3) Dynamic Selection}: dynamic selection maximizes the performance of the model and reduces unnecessary information redundancy by selecting the most useful feature channels for exchange. The exchange ratio is judiciously set to 0.5 to maintain a balance.

As shown in Table~\ref{tab:DynamicExchange}, the advantages of dynamic selection are obvious. Empirical evidence highlights the robustness and optimality of dynamic selection, thus confirming its adoption as the preferred method for feature selection in our model.

Additionally, we conduct a focused analysis on the impact of the hyperparameter $p$ in the Top-K selection mechanism. As shown in Table~\ref{tab:Top-K}, setting $p = 1/2$ yields the best performance, validating this choice as optimal for our design.

\subsubsection{\textbf{Impact of Backbone Depth}}
In infrared small target detection, the depth of the backbone network plays a pivotal role in determining detection performance. A shallower backbone, though computationally efficient, may fail to capture sufficiently rich semantic features, thereby limiting the model’s ability to distinguish small targets from complex backgrounds. On the other hand, an overly deep backbone may introduce excessive abstraction, potentially suppressing the subtle features of small targets and leading to performance degradation. Therefore, achieving an optimal balance between depth and feature representation is essential for effective infrared target detection.


We conducted a series of experiments using different ResNet architectures, varying the depth from ResNet18 to ResNet101, to identify the optimal backbone for our proposed BAFE-Net. The results of these experiments are summarized in Table~\ref{tab:Resnet}.
Notably, on DenseSIRST, ResNet18 has the highest $mAP_{07}$ metric and ResNet50 has the highest $mAP_{12}$ metric; on IRSTD-1k, ResNet18 has the highest $mAP_{07}$ metric and ResNet34 has the highest $mAP_{12}$ metric.



To further validate the effectiveness of our proposed methods, we conduct a visual analysis by comparing the feature activation maps of our approach with those of the top-performing competing methods in the quantitative evaluation. Fig.~\ref{fig:heatmap} illustrates the feature activation maps of the five most competitive methods, including our own, on representative samples from the DenseSIRST dataset.

From the visual results, we observe that our method exhibits two key advantages over the other approaches. First, the feature activation maps of our method demonstrate significantly stronger activation intensities in the target regions, indicating a higher sensitivity to the presence of infrared small targets. The deeper activation levels in the target areas suggest that our method is more effective in capturing the discriminative features of the targets, contributing to improved detection accuracy.

Second, our method generates cleaner background activation maps. As evident from Fig.~\ref{fig:heatmap}, the background regions in our activation maps exhibit consistently lower activation intensities, effectively suppressing background interference. This can be attributed to the background-aware mechanism employed by BAFE-Net, which learns to differentiate between target and background regions during the feature exchange process. By minimizing the activation levels in the background areas, our method demonstrates enhanced robustness against false positives, leading to a higher detection precision.


\section{Conclusion} \label{sec:conclusion}


This study tackles the key challenges in infrared small target detection by leveraging background semantics.
We introduce DenseSIRST, a benchmark dataset with detailed background annotations, enabling the transition from sparse to dense target detection.
Building upon this, we propose BAFE-Net, a multi-task architecture that explicitly models background semantics and facilitates cross-task feature hard-exchange, promoting the exchange of information between the classification and background segmentation branches.
Moreover, we develop BAG-CP, a synthesis method for the DenseSIRST dataset, which also serves as a data augmentation technique. BAG-CP selectively pastes targets into sky regions, outperforming the original Copy-Paste method. 
Extensive experiments validate the effectiveness of BAFE-Net in improving detection accuracy and reducing false alarms. 
Our work underscores the importance of explicitly leveraging contextual information in infrared small target detection and paves the way for the development of more robust and context-aware algorithms in this domain.

\bibliographystyle{IEEEtran}
\bibliography{./reference.bib}


\end{document}